\documentclass{article}

\makeatletter
\def\ps@headings{%
	\def\@oddhead{\mbox{}\scriptsize\rightmark \hfil \thepage}%
	\def\@evenhead{\scriptsize\thepage \hfil \leftmark\mbox{}}%
	\def\@oddfoot{}%
	\def\@evenfoot{}}
\makeatother
\usepackage[numbers,sort &compress]{natbib}
\usepackage[final]{nips_2018}
\usepackage{algorithm,algorithmic}
\usepackage{amsfonts,amsmath,amssymb,amsthm}
\usepackage{bm}
\usepackage{booktabs}       
\usepackage{color}
\usepackage{enumitem}
\usepackage{epsfig}
\usepackage{geometry,graphicx}
\usepackage{hyperref}       
\usepackage{mathrsfs}
\usepackage{microtype}      
\usepackage{multicol,multirow}
\usepackage{setspace}
\usepackage{subcaption}
\usepackage{url}   

\title{Deep Reinforcement Learning for Intelligent Transportation Systems}

\author{%
  Xiao-Yang Liu \\
  Department of Electrical Engineering\\
  Columbia University\\
  \texttt{XL2427@columbia.edu} \\
   \And
  Zihan Ding\\
  Department of Computing\\
  Imperial College London\\
    \texttt{zd2418@ic.ac.uk} \\
   \AND
     Sem Borst\\
   Nokia-Bell Laboratories\\
    \texttt{sem.borst@nokia-bell-labs.com} \\
   \And
    Anwar Walid\\
   Nokia-Bell Laboratories\\
    \texttt{anwar.walid@nokia-bell-labs.com} \\
}

\begin{document}

\maketitle

\begin{spacing}{1.0}

\begin{abstract}
Intelligent Transportation Systems (ITSs) are envisioned to play a critical role in improving traffic flow and reducing congestion, which is a pervasive issue impacting urban areas around the globe. Rapidly advancing vehicular communication and edge cloud computation technologies provide key enablers for smart traffic management. However,
operating viable real-time actuation mechanisms on a practically relevant scale involves formidable challenges, e.g., policy iteration and conventional
Reinforcement Learning (RL) techniques suffer from poor scalability due to state space explosion. Motivated by these issues, we explore the potential for Deep Q-Networks
(DQN) to optimize traffic light control policies. As an initial benchmark, we establish that the DQN algorithms yield the ``thresholding" policy in a single-intersection.
Next, we examine the scalability properties of DQN algorithms and their performance in a linear network topology with several intersections along a main artery. We demonstrate that DQN algorithms produce intelligent behavior, such as the emergence of ``greenwave'' patterns, reflecting their ability to learn favorable traffic light actuations.
		
		
\end{abstract}

\section{Introduction}
\label{intr}

Emerging Intelligent Transportation Systems (ITSs) \cite{vazifeh2018addressing, zhu2016public, zhu2018online, zhu2018joint, alam2016introduction,lv2015traffic,sundar2015implementing} are expected to play
an instrumental role in improving traffic flow,
thus optimizing fuel efficiency, reducing delays and enhancing the
overall driving experience.
Today traffic congestion is an exceedingly complex and vexing issue
faced by metropolitan areas around the world.
In particular, street intersections in dense urban traffic zones
(e.g.,~Times Square in Manhattan) can act as severe bottlenecks.

Current traffic light control policies typically involve preprogrammed
cycles that may be optimized based on historical data and adapted
according to daily patterns.
The options for adaptation to real-time conditions, e.g.~through
detection wires in the pavement, tend to be fairly rudimentary.
Evolving vehicular communication technologies offer a crucial
capability to obtain more fine-grained knowledge of the positions
and speeds of vehicles.
Such comprehensive real-time information can be leveraged,
in conjunction with edge cloud computation, for significantly improving
traffic flow through more agile traffic light control policies,
or in the longer term, via direct actuation instructions for fully
automated driving scenarios \cite{qi2018two}.
While the potential benefits are immense, so are the technical
challenges that evidently arise in solving such real-time actuation
problems on an unprecedented scale in terms of intrinsic
complexity, geographic range, and number of objects involved.

Under suitable assumptions, the problem of optimal dynamic traffic
light control may be formulated as a Markov decision process (MDP)
\cite{onori2016dynamic,puterman2014markov,ross2014introduction}.
The MDP framework provides a rigorous notion of optimality along
with a basis for computational techniques such as value iteration,
policy iteration \cite{toolbox} or linear programming.
However, methods like policy iteration involve strong model assumptions,
which may not always be satisfied in reality, and knowledge
of relevant system parameters, which may not be readily available.
Owing to these issues, the policy iteration approach tends to be vulnerable
to model mis-specification and inaccurate parameter estimation.
Moreover, in terms of computational aspects, the policy iteration approach suffers
from the curse of dimensionality, resulting in excessively large state
spaces in realistic problem instances and exceedingly slow convergence.


Reinforcement Learning (RL) techniques, such as Q-learning, overcome
some of these limitations \cite{barto1991real,moore1993prioritized,sutton1988learning,watkins1992machine} 
and have been previously considered in the context of optimal dynamic
traffic light control
\cite{van2016video,van2016coordinated,wiering2000multi, wei2018intellilight}.
However, conventional RL techniques are still prone to prohibitively
large state spaces and extremely sluggish convergence,
implying poor scalability beyond a single-intersection scenario.

Motivated by the above issues, we explore in the present paper the
potential for deep learning algorithms, particularly Deep Q-Networks (DQN)~\cite{mnih2015human}, to optimize real-time traffic light control policies in large-scale transportation systems.
As an initial validation benchmark, we analyze a single-intersection
scenario and corroborate that the DQN algorithms match the provably
optimal performance achieved by the policy iteration approach and exhibit
a similar threshold structure.
Next, we consider a linear network topology with several intersections
to examine the scalability properties of DQN algorithms
and their performance in the presence of highly complex interactions
created by the flow of vehicles along the main artery.
As mentioned above, the use of the policy iteration approach or standard RL techniques
involves an excessive computational burden in these scenarios;
hence the optimal achievable performance cannot be easily quantified.
As a relevant qualitative feature, we demonstrate that DQN algorithms
produce intelligent behavior, such as the emergence of ``greenwave''
patterns \cite{lammer2008self,lammer2010self}, even though such structural
features are not explicitly prescribed in the optimization process.
This emergent intelligence confirms the capability of the DQN algorithms
to learn favorable structural properties solely from observations.

The remainder of the paper is organized as follows.
In Section~\ref{mode}, we present a detailed model description
and problem statement.
In Section~\ref{algo}, we provide a specification of the DQN algorithms
for a single intersection as well as a linear network with several
intersections.
Section~\ref{perf} discusses the computational experiments conducted
to evaluate the performance of the proposed DQN algorithms
and illustrate the emergence of ``greenwave'' patterns.
In Section~\ref{conc}, we conclude with a few brief remarks and some
suggestions for further research.

\section{Model Description and Problem Statement}
\label{mode}

We model the road intersections and formulate our optimization problem. For the sake of transparency, we consider an admittedly stylized model that only aims to capture the most essential features that govern the dynamics of contending traffic flows at road intersections.
We throughout adopt a discrete-time formulation to simplify the description and allow direct application of MDP techniques for comparison, but the methods and results naturally extend to continuous-time operation.

\begin{figure}[t]
	\centering
	\includegraphics[width=0.3\textwidth]{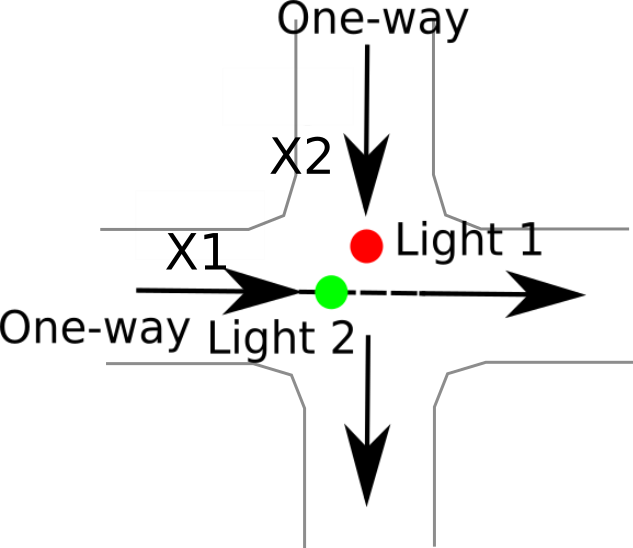}
	\hspace*{0.15\textwidth}
	\includegraphics[width=0.4\textwidth]{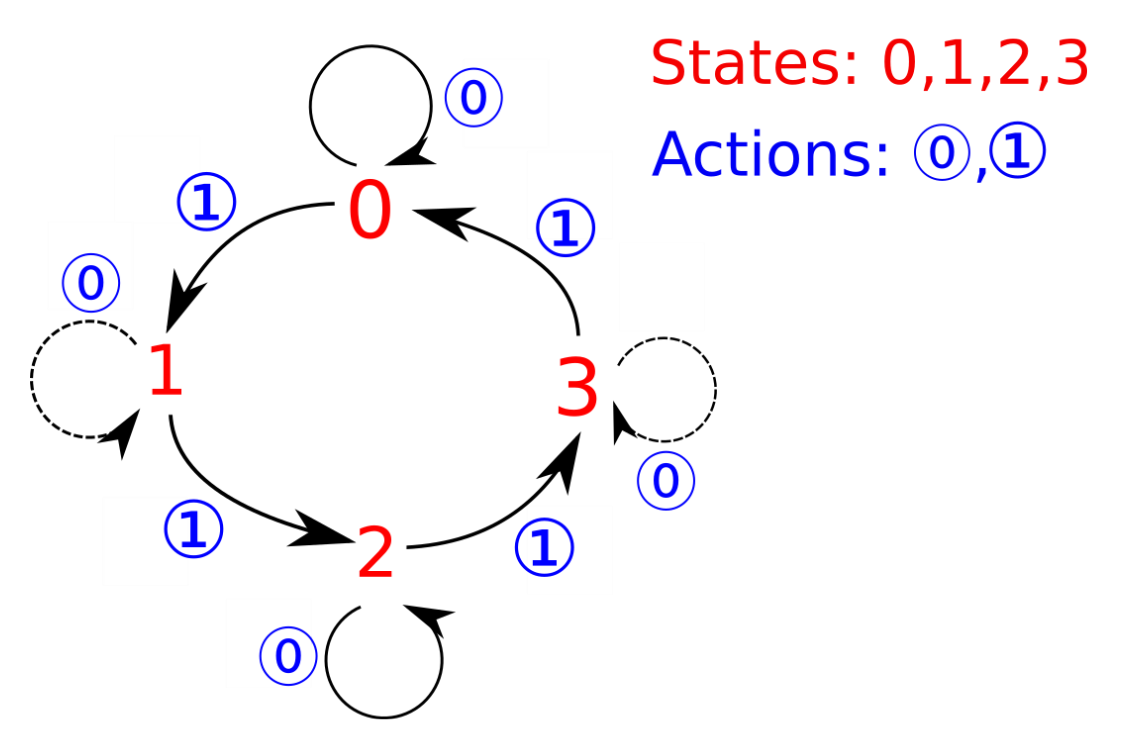}
	\caption{One intersection with two traffic flows (left) where $X_1$ and $X_2$ are the queue lengths, and the state transition diagram (right).}
	\label{fig:logo}
	\vspace{-0.2in}
\end{figure}

\begin{figure}[b]
	\centering
	\includegraphics[width=0.70\textwidth]{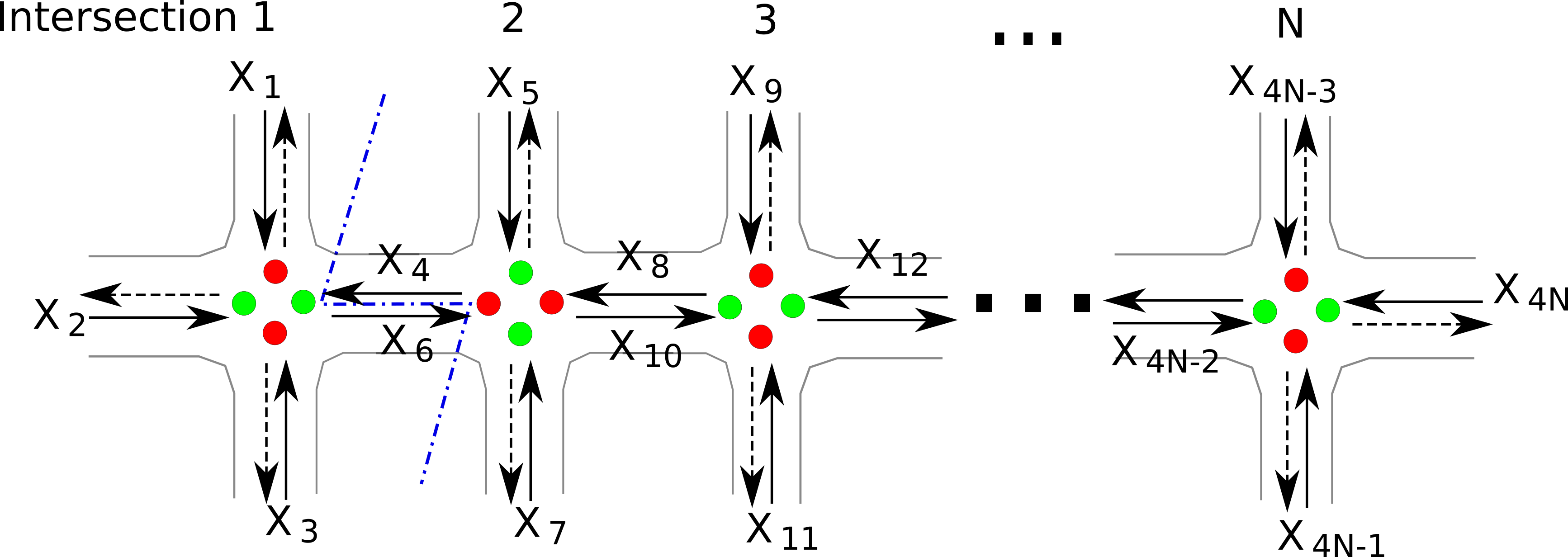}
	\hspace*{0.04\textwidth}
	\caption{Linear bidirectional road network.} 
	\label{fig:line}
	\vspace{-0.2in}
\end{figure}

\subsection{Single Road Intersection}
\label{modesingle}

As mentioned earlier, we start with a single-intersection scenario
to facilitate the validation of the DQN algorithms by comparing it
with the policy iteration approach.
We consider the simplest meaningful setup with two intersecting
unidirectional traffic flows as schematically depicted in the left
side of Fig.~\ref{fig:logo}.
The state $S(t)$ of the system at the beginning of time slot~$t$
may be described by the three-tuple $(X_1(t), X_2(t); Y(t))$,
with $X_i(t)$ denoting the number of vehicles of traffic flow~$i$
waiting to cross the intersection and $Y(t) \in \{0, 1, 2, 3\}$
indicating the configuration of the traffic lights:
\begin{itemize}
	\item ``0":
	green light for direction~$1$ and hence red light for direction~$2$;
	\item ``1":
	yellow light for direction~$1$ and hence red light for direction~$2$;
	\item ``2":
	green light for direction~$2$ and hence red light for direction~$1$;
	\item ``3":
	yellow light for direction~$2$ and hence red light for direction~$1$.
\end{itemize}

Each configuration~$k$ can either simply be continued in the next
time slot or must otherwise be switched to the natural subsequent
configuration $(k + 1) \mod 4$.
This is determined by the action $A(t)$ selected at the end of time
slot~$t$, which is represented by a binary variable as follows:
``0" for continue, and ``1" for switch:
\begin{equation}
Y(t+1) = (Y(t) + A(t)) \mod 4.
\end{equation}
These rules give rise to a strictly cyclic control sequence
as illustrated in the right side of Fig.~\ref{fig:logo}.



The evolution of the queue state over time is governed by the recursion
\begin{equation}
(X_1(t+1), X_2(t+1)) =
(X_1(t) + C_1(t) - D_1(t), X_2(t) + C_2(t) - D_2(t)),
\label{lindleysingle}
\end{equation}
with $C_i(t)$ denoting the number of vehicles of traffic flow~$i$
appearing at the intersection during time slot~$t$
and $D_i(t)$ denoting the number of departing vehicles
of traffic flow~$i$ crossing the intersection during time slot~$t$.
While not essential for our analysis, we make the simplifying
assumption that if one of the two traffic flows is granted the
green light, then exactly one waiting vehicle of that traffic flow,
if any, will cross the intersection during that time slot, i.e.,
\begin{equation}
D_1(t) = \min\{1, X_1(t)\} \mbox{ if } Y(t) = 0;
D_2(t) = \min\{1, X_2(t)\} \mbox{ if } Y(t) = 2;
\label{servicesingle}
\end{equation}
and $D_1(t) = 0$ if $Y(t) \neq 0$ and $D_2(t) = 0$ if $Y(t) \neq 2$.

\subsection{Linear Road Topology}
\label{modemultiple}


To examine the performance and scalability properties of the DQN algorithms in more complex large-scale scenarios, we will consider a linear road topology. 
Specifically, we investigate a linear network topology
with $N$~intersections and bidirectional traffic flows,
representing a main artery with cross streets as schematically
depicted in Fig~\ref{fig:line}.
We do not account for any traffic flows making left or right turns, but the analysis could easily be generalized to accommodate that.
The state $S(t)$ of the system at the beginning of time slot~$t$ may
be described by a $(5 N)$-tuple
$(X_{n1}(t), X_{n2}(t), X_{n3}(t), X_{n4}(t); Y_n(t))_{n = 1, \dots, N}$,
with directions~$1$ and~$2$ corresponding to the east-west direction of the main artery and the north-south direction of the cross streets,
and thus
\begin{equation}
Y_n(t+1) = (Y_n(t) + A_n(t)) \mod 4,
\label{actionsingle}
\end{equation}
with $A_n(t)$ denoting the action selected for the $n$-th
intersection at the end of time slot~$t$.


The evolution of the various queue states is governed by the recursion
\begin{equation}
X_{ni}(t+1) = X_{ni}(t) + C_{ni}(t) - D_{ni}(t),
\label{lindleymultiple}
\end{equation}
with $C_{ni}(t)$ denoting the number of vehicles in direction~$i$
appearing at the $n$ th intersection during time slot~$t$ and $D_{ni}(t)$ denoting the number
of vehicles crossing the $n$-th intersection in direction~$i$
during time slot~$t$, $i = 1, \dots, 4$, $n = 1, \dots, N$.
While $C_{11}(t)$, $C_{N3}(t)$, and $C_{n2}(t)$, $C_{n4}(t)$,
$n = 1, \dots, N$, correspond to vehicles approaching the intersection
from the external environment, we have $C_{n+1,1}(t+u) = D_{n1}(t)$
and $C_{n3}(t+u) = D_{n+1,3}(t)$, $n = 1, \dots, N -1$.
This reflects that the vehicles crossing the $n$-th intersection
in eastern direction during time slot~$t$ appear at the $(n+1)$-th
intersection $u$~time slots later; likewise, vehicles passing
through the $(n+1)$-th intersection in western direction during time
slot~$t$ arrive at the $n$-th intersection $u$~time slots later.
In this manner, the vehicles that travel along the main artery create
highly complex interactions among the various intersections,
which present additional challenges in optimizing the control policy.

Note that
\begin{equation}
D_{n1}(t) = \min\{1, X_{n1}(t)\}, D_{n3}(t) = \min\{1, X_{n3}(t)\}, 
\mbox{ if } Y_n(t) = 0,
\label{servicemultiple}
\end{equation}
$D_{n1}(t), D_{n3}(t) = 0$ if $Y_n(t) \neq 0$, and similarly for
$D_{n2}(t)$ and $D_{n4}(t)$ depending on whether $Y_n(t) = 2$ or not.

\subsection{Optimization Goal}

We assume that the ``congestion cost'' in time slot~$t$ may be expressed
as a function $F(X(t))$ of the queue state,
with $X(t) = (X_1(t), X_2(t))$ in the single-intersection scenario
and $X(t) = (X_{n1}(t), X_{n2}(t), X_{n3}(t), X_{n4}(t))_{n = 1, \dots, N}$
in the linear topology with $N$~intersections.
The goal is to find a dynamic control policy which selects actions
over time so as to minimize the long-term expected discounted cost
$\mathbb{E}\left[\sum_{t=1}^{\infty} \gamma^t F(X(t))\right]$,
with $\gamma \in (0, 1)$ representing a discount factor.

\section{Algorithm Design}
\label{algo}

We provide a detailed specification of the DQN algorithms
for the scenarios of a single intersection or a linear topology
with several intersections as described in the previous section.

First of all, let $Q(s, a)$ be the maximum achievable expected
discounted reward (or minimum negative congestion cost in our context)
under the optimal policy starting from state $s = (X; Y)$
when action~$a$ is taken.
The $Q(s, a)$ values satisfy the equations
\begin{equation}
Q(s, a) = r(s, a) + \gamma \sum\limits_{s' \in \mathcal{S}}
p(s, s'; a) \underset{a' \in \mathcal{A}}\max Q(s', a') = r(s, a) +
\gamma \mathbb{E}\left[\underset{a' \in \mathcal{A}}\max Q(s', a')\right],
\label{qvalues}
\end{equation}
with $r(s, a) = F(X)$ denoting the congestion cost in queue state~$X$,
and $p(s, s'; a)$ denoting the transition probability from state~$s$
to state~$s'$ when action~$a$ is taken.
Observe that the values $V(s) = \underset{a' \in \mathscr{A}} \max Q(s, a')$
satisfy the Bellman optimality equations
\begin{equation}
V(s) = \underset{a' \in \mathcal{A}}\max \{r(s, a) +
\gamma \sum\limits_{s' \in \mathcal{S}} p(s, s'; a) V(s') \} =
\underset{a' \in \mathcal{A}}\max \{r(s, a) + \gamma \mathbb{E}[V(s')]\}.
\label{bellman}
\end{equation}

The system state $S$ serves as the input for both the target network
and the evaluate network in the DQN algorithms,
with $S = (X_1, X_2; Y)$ in the single-intersection scenario
and $S = (X_{n1}, X_{n2}, X_{n3}, X_{n4}; Y_n)_{n = 1, \dots, N}$
in the linear topology with $N$~intersections.
Equation~\eqref{qvalues} provides the basis for deriving the target
Q-values at each time step, while the Q-learning update for the neural
network approximator in the $i$-th iteration is calculated based on
\begin{equation}\label{problem_obj_function}
\text{Loss}(\theta_i) = \mathbb{E}_{s,a,r,s'\sim memory}\bigg[\bigg(r+\gamma\max\big( q\_target(s',a';\theta'_i)\big)-q\_eval(s,a;\theta_i)\bigg)^2\bigg],
\end{equation}
where $r$ is reward (negative cost) in the current step,
$s'$ and $a'$ are the state and action in the next step,
$\theta_i$ are parameters of the evaluate Q-network in the $i$-th
iteration and $\theta'_i$ are parameters of the target Q-network
with delayed update following the evaluate network.

The DQN algorithms sample from and train on data collected in memory.
The online samples are stored in memory for further learning.
A warm-up period of $k_0$ time steps is applied before the learning
operations are initiated.
The evaluate network is updated with the AdamOptimizer \cite{kingma2014adam}
gradient-descent and $\epsilon$-greedy policy, whereas the update of the target network is slightly later.

Based on the above outline, we provide the specification
of the DQN algorithm for the single-intersection
scenario Fig. 1 and a linear topology
with $N$~intersections in Fig. 2.
It is worth observing that even in the latter case we adopt
a ``single-agent'' DQN algorithm which has access to the global state of the network, as opposed to the ``multiple-agent'' method with one agent for each individual intersection as considered in \cite{satunin2014multi,wiering2000multi}.
While the single-agent approach involves a larger state space, it allows more intelligent control and coordination on a global level, which manifests itself for example in the emergence of greenwave patterns as we will demonstrate in the next section.

\begin{algorithm}[t]
	\caption{DQN for single intersection or linear road topology
		with $N$~intersections}
	\label{alg_SAMI}
	\begin{algorithmic}
		\STATE 1:~Initialize queue and control states:
		either $X_1, X_2 = 0; Y = 0$ [single intersection]
		or $X_{n1}, X_{n2}, X_{n3}, X_{n4} = 0; Y_n = 0$ for all $n = 1, \dots, N$
		[linear topology];
		\STATE 2:~For steps $k = 1, \dots, K$ do:
		\STATE 3:~~~~~~$s = [X_1, X_2; Y]$ [single intersection]
		or $s = [X_{11}, \dots, X_{N4}; Y_1, \dots Y_N]$
		[linear topology];
		\STATE 4:~~~~~Select action $a$ with
		$a^* = \arg\underset{a \in \mathcal{A}}\max~ q\_{eval}(s; a)$ and $\epsilon$-greedy policy,
		using eval\_net to evaluate the $Q$-value for each action;
		\STATE 5:~~~~~Generate random variables $C_1$, $C_2$ [single intersection]
		or $C_{11}$, $C_{N3}$ and $C_{n2}$, $C_{n4}$ for all $n = 1, \dots, N$;
		\STATE 6:~~~~~~Given~$a$, determine new queue and control states
		$X_1', X_2', Y'$
		according to Eq.~(1)-(3)
		[single intersection]
		or $X_{11}', \dots, X_{N4}', Y_1', \dots, Y_N'$ 
		according to Eq.~(4)-(6)
		[linear topology];
		\STATE 7:~~~~~~$r = - ((X_1')^2 + (X_2')^2)$ [single intersection]
		or $r = - \sum_{n=1}^{N} \sum_{i = 1}^{4} (X_{ni}')^2$ [linear topology];
		\STATE 8:~~~~~~$s' = [X_1', X_2'; Y']$ [single intersection]
		or $s' = [X_{11}',\dots, X_{N4}'; Y_1', \dots, Y_N']$
		[linear topology];
		\STATE 9:~~~~~Store transition $[s, a, r, s']$ in memory;
		\STATE 10:~~~~Perform learning operation if $k > k_0$:
		\STATE 11:~~~~~~~~~~Sample a minibatch of samples from memory;
		\STATE 12:~~~~~~~~~~Update target network: $\theta'_i=\theta_i$;
		\STATE 13:~~~~~~~~~~Calculate the target Q-value:
		$q\_target(s, a) = r +
		\gamma \underset{a' \in \mathcal{A}}\max q\_target(s', a')$;
		\STATE 14:~~~~~~~~~~Update evaluate network with gradient descent
		(using AdamOptimizer):
		$\text{Loss}(\theta_i)=\mathbb{E}[(q\_target-q\_eval)^2]$.
	\end{algorithmic}
	\vspace{-2pt}
\end{algorithm}

\section{Performance Evaluation}
\label{perf}

We present simulations
to evaluate the performance of the DQN algorithm in Alg. 1, and in particular illustrate the emergence
of ``greenwave'' patterns in linear topology networks. Our codes are available at \cite{codes}.

\subsection{Single Road Intersection}
\label{resusingle}

As an initial validation benchmark, we first consider
a single-intersection scenario as described in Subsection~\ref{modesingle}.
The reason for considering this toy scenario is that the state space
is sufficiently small for the optimal policy to be computed using
the baseline policy iteration approach.
We assume the numbers of arriving vehicles of both traffic flows
in each time step as represented by the random variables~$C_1$ and~$C_2$
to be independent and Bernoulli distributed with parameter $p = 1/4$.
We use a quadratic congestion cost function
$F(X_1, X_2) = X_1^2 + X_2^2$ and a discount factor $\gamma = 0.99$. 

\begin{figure}[t]
	\includegraphics[width=0.46\textwidth]{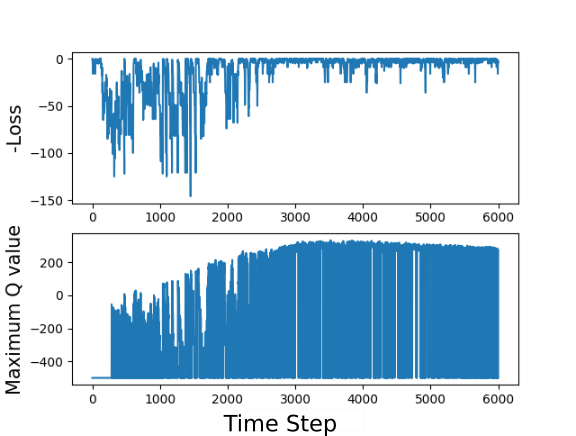}
	\hspace*{0.04\textwidth}
	\includegraphics[width=0.46\textwidth]{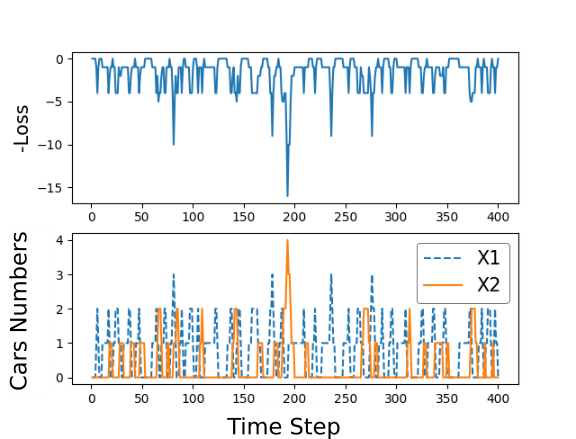}
	\caption{Learning curve for DQN policy (left) and thresholding
		property of DQN policy (right).}
	\label{single_intersection_results}
	\vspace{-0.2in}
\end{figure}

Inspection of the results in Fig. \ref{single_intersection_results} shows that the DQN policy as obtained
using Alg.~1 coincides with the optimal policy with the traditional policy iteration method.
In particular, it matches the optimal performance and exhibits
a similar threshold structure.
This structural property was also reported in~\cite{hofri1987optimal}
for a strongly related two-queue dynamic optimization problem
(with switch-over costs rather than switch-over times).

\begin{figure}[t]
	\includegraphics[width=0.65\textwidth]{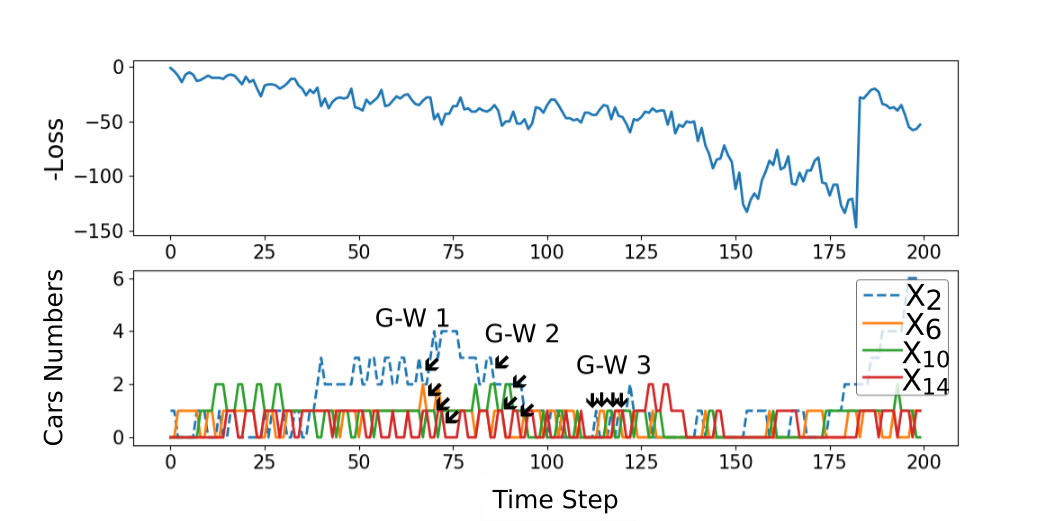}
	\hspace*{0.002\textwidth}
	\includegraphics[width=0.37\textwidth]{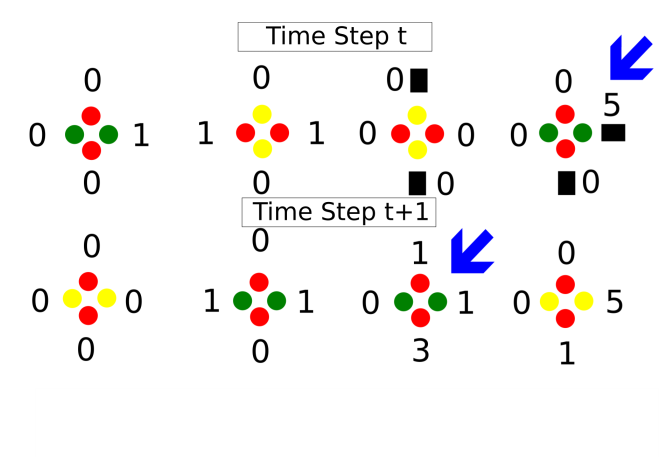}
	\caption{Testing results for the linear network of size $4\times1$ (left) and Greenwave traffic lights (right). 
		``Greenwave'' is abbreviated to be ``G-W''. In the right, numbers indicate the number of vehicles waiting on each road. Black rectangles indicate incoming vehicles from peripheral roads.
	}
	\vspace{-0.2in}
\end{figure}

\subsection{Linear Road Topology}
\label{resumultiple}

We now turn to the scenario
in Fig. 2.
This is a more challenging scenario which serves to examine the
scalability properties of our algorithm and its performance in the
presence of highly complex interactions arising from the flow
of vehicles along the main east-west arterial road.

Assume the numbers of externally arriving vehicles in eastern and western directions 
in each time step, represented by the random variables $C_{11}$ and $C_{N3}$, to be independent and Bernoulli distributed with parameter~$p_1=1/4$.
The numbers of arriving vehicles in southern and northern directions on each of the $N$~cross streets in each time step, represented by the random variables $C_{n2}$ and $C_{n4}$, $n = 1, \dots, N$, are  also independent and Bernoulli distributed with parameter~$p_2=1/8$.
We use a quadratic congestion cost function
$F(X) = \sum_{n = 1}^{N} \sum_{i = 1}^{4} X_{ni}^2$ and a discount
factor $\gamma = 0.99$. In simulations, the evaluate and target networks used in Alg. 1 have both $4$ fully-connected layers of size $200, 100, 40$ and $2$, respectively. We use ReLu as activation functions and squared difference loss.

The use of a policy iteration approach is computationally infeasible in this case
due to the state space explosion, and hence the degree of optimality
of our algorithm cannot be assessed in a quantitative manner.
Instead we have therefore examined qualitative features to validate the
intelligent behavior of our algorithm and evaluate its performance merit.
In particular, we observed the emergence of ``greenwave'' patterns as shown in Fig. 4, even though such structural features are not explicitly prescribed
in the optimization process. Specifically, the ``greenwave'' phenomenon is reflected as consecutive reduction of car numbers in each road.
This emergent intelligence confirms the capability of our algorithm
to learn favorable structural properties solely from observations.

\section{Conclusion}
\label{conc}

We have explored the scope for Deep Q-Networks (DQN) to optimize
real-time traffic light control policies in emerging large-scale
Intelligent Transportation Systems. As an initial benchmark, we established that DQN algorithms deliver the optimal performance achieved by the policy iteration approach
in a single-intersection scenario. We subsequently evaluated the scalability properties of DQN algorithms in a linear topology with several intersections, and demonstrated the emergence of intelligent behavior such as ``greenwave'' patterns, confirming their ability to learn desirable structural features. 

In future research we intend to investigate locality properties and analyze how these can be exploited in the design of distributed coordination schemes for wide-scale deployment scenarios. It would be interesting to investigate the effectiveness of other RL methods, like Deep Deterministic Policy Gradients (DDPG) used in \cite{Xiong2018}, for transportation systems.
		
	
	
	
\medskip
	
\small

\bibliographystyle{plain}
\bibliography{ref}
	
\end{spacing}
	
\end{document}